\pdfoutput=1

\documentclass[11pt]{article}

\usepackage[preprint]{acl}

\usepackage{times}
\usepackage{latexsym}
\usepackage{colortbl}
\usepackage{xcolor}
\usepackage{arydshln}
\usepackage{comment}
\usepackage{booktabs}
\usepackage{float}

\usepackage{multirow}

\definecolor{darkgreen}{rgb}{0.0,0.5,0.0}
\usepackage[T1]{fontenc}

\usepackage[utf8]{inputenc}

\usepackage{microtype}

\usepackage{inconsolata}

\usepackage{graphicx}

\title{Personalized Author Obfuscation with Large Language Models}

\author{Mohammad Shokri \\
  The Graduate Center \\
  CUNY \\
  \texttt{mshokri@gradcenter.cuny.edu} \\\And
  Sarah Ita Levitan \\
  Hunter College \\
  CUNY \\\And
  Rivka Levitan \\
  Brooklyn College \\
  CUNY\\}

\begin{document}
\maketitle
\begin{abstract}
In this paper, we investigate the efficacy of large language models (LLMs) in obfuscating authorship by paraphrasing and altering writing styles. Rather than adopting a holistic approach that evaluates performance across the entire dataset, we focus on user-wise performance to analyze how obfuscation effectiveness varies across individual authors. While LLMs are generally effective, we observe a bimodal distribution of efficacy, with performance varying significantly across users. To address this, we propose a personalized prompting method that outperforms standard prompting techniques and partially mitigates the bimodality issue.
\end{abstract}

\section{Introduction}
Author Attribution (AA) and Author Verification (AV) are two classic problems in Natural Language Processing. AA involves predicting the author of a text \textit{T} from a set of users. AV is a specific case of AA where we verify whether an author $u_i$ is the writer of a given \textit{T}, turning it into a binary classification problem. With the abundance of online data and advancements in transformer-based language models, AA and AV have become easier tasks than ever. The emergent power of LLMs poses significant privacy threats \cite{staab2023beyond}, particularly to  journalists and human rights activists working under authoritarian regimes who could be affected by successful AA and AV attacks. 

To defend against these models, researchers propose employing \textit{Author obfuscation (AO)} techniques to anonymize their writing by changing their writing style while retaining the meaning of the text. With the rise of ChatGPT and similar models and their fast acceptance in the world, the standard for fluency in algorithm-generated text has increased, making rigid rule-based methods less appealing to users~\cite{fisher2024jamdec}. These widely accessible models are likely to be used for AO by vulnerable authors, making it crucial to assess their effectiveness for this purpose. Recent research highlights paraphrasing as a robust author obfuscation method~\cite{tripto2023ship,fisher2024styleremix,bevendorff2019heuristic,almishari2014fighting}. Consequently, large language models (LLM) have been examined as a natural solution and have been demonstrated to have strong obfuscation performance~\cite{mattern2022limits,utpala2023locally, fisher2024jamdec}. Despite reporting excellent performance, most studies report broad statistics on obfuscation performance across multiple datasets, limiting our understanding of how effective the obfuscation is for individual users~\cite{utpala2023locally, fisher2024jamdec}. 

Our goal in this study is to explore user-level inconsistencies in LLM paraphrasing and analyze how these variations manifest across individuals. We aim to determine whether such inconsistencies can be leveraged to develop a personalized paraphrasing approach by exploiting the abundance of user data. Our research questions in this study are as follows:

\begin{itemize}
    \setlength{\itemsep}{0pt}
    \setlength{\parskip}{0pt}
    \item RQ1. How effectively can LLMs evade authorship detection?
    \item RQ2. How effectively can LLM paraphrasing be tailored for personalized obfuscation?
\end{itemize}

In this paper, we attempt to answer these questions using GPT-4~\cite{achiam2023gpt} (\texttt{gpt-4-turbo}) and LLaMA-3.1 \cite{dubey2024llama} (\texttt{meta-llama/Llama-3.1-8B-Instruct}), two widely adopted and powerful LLMs in different sizes. We explore prompt-based paraphrasing and its performance consistency across different users in a zero-shot setting, where we simply ask the model to paraphrase the text to hide its author's identity.  In addition, motivated by the observed performance variability across authors, we examine the potential of personalized prompting based on key writing style features unique to each author. We use SHAP values \cite{hart1989shapley} to identify these key features for each author separately, and design a user-specific prompt to tweak the identified feature while paraphrasing.

\section{Related Work}
Early AO studies used rule-based methods for  sentence transformations, such as contraction replacement or synonym substitution \cite{castro2017author, karadzhov2017case, potthast2016author}. These methods are simple and fast, but reduce fluency and semantic similarity.
Many researchers treat AO as a an adversarial attack on  AA/AV models, aiming to minimally perturb the input to ensure misclassification while maintaining semantic similarity \cite{gao2018black, ebrahimi2017hotflip}. However, adversarial perturbations often degrade text quality~\cite{crothers2022adversarial}. 

In order to change the writing style, some studies explored re-writing methods such as back translations \cite{keswani2016author, altakrori2022multifaceted, bo2019er}. Although effective, these approaches can produce unnatural phrasing and semantic loss. Variational auto-encoders and generative adversarial networks have also been explored for obfuscation~\cite{shetty2018a4nt, mireshghallah2021style}. Mutant-X~\cite{mahmood2019girl} and Avengers \cite{haroon2021avengers} use a genetic algorithm to iteratively substitute words until the text fools the internal classifier. Alison \cite{xing2024alison} is a faster syntactical AO method which replaces multi-token phrases to fool an internal classifier trained on character and POS n-grams.

Differential Privacy approaches add noise to the vector representation of input at the word-level and sentence-level \cite{feyisetan2020privacy, meehan2022sentence,mattern2022limits}. DP-Prompt~\cite{utpala2023locally} integrates differential privacy perturbation with paraphrasing to generate tokens in the paraphrased document under a privacy preserving framework. \citet{tripto2023ship} studies the effect of a sequence of LLM paraphrasing on a given text and find that LLMs impose their distinctive style onto the paraphrased text, and that they generally preserve content pretty well. Most similar to our work, is StyleRemix~\cite{fisher2024styleremix} where the authors propose an interpretable personalized obfuscation method based on style elements, limited to seven predefined axes. Their method uses LoRA modules to modify writing style across seven axes: formality, length, sentiment, complexity, concreteness, directness, and narrativity. Unlike previous studies, our work investigates variation of performance with focus on individuals, and present a personalized prompting solution that leverages a broader range of style elements.
\section{Data}

We work with data samples from three datasets to ensure generalizabilty of our results across different domains. We evaluate the performance of our authorship obfuscation approach on these three widely used datasets, which are relatively large in terms of the total number of reviews and posts per user.

\textbf{IMDb.} The IMDB62 dataset \cite{seroussi2014authorship} consists of user reviews from the Internet Movie Database (IMDb). It contains 62,000 reviews written by 62 distinct authors, with approximately 1,000 reviews per author. The dataset is widely used in authorship attribution tasks because it provides a balanced and relatively clean source of personal writing. The reviews are highly subjective, often featuring personal opinions and informal language, making the dataset useful for evaluating models' ability to capture nuanced stylistic differences among authors. We select the 10 users used in DP-Prompt \cite{utpala2023locally} to work with in our study. Each user has 1350 reviews in our data. We split the data into 80\% training, 10\% validation, and 10\% test sets. The average word count per review is 234 words, making it the longest on average among the datasets used in this study.

\textbf{Yelp.} The second dataset is the Yelp dataset on Github:https://github.com/sixhobbits/yelp-dataset-2017/tree/master. This dataset contains a wide range of writing styles and linguistic patterns across different domains such as restaurants, services, and businesses. It provides a rich source for authorship analysis, as the reviews vary in length, sentiment, and content, allowing for the exploration of stylistic differences between authors. The dataset is commonly used in authorship attribution and verification tasks due to its diverse set of users and high variability in writing style. From the 45 available users, we randomly select 10 users, each with approximately 500 posts. The average word count per review in this dataset is 173 words. We split the data for each user into 80\% training, 10\% validation, and 10\% test sets.

\textbf{Blog.}
The third dataset consists of diary-style blog posts \cite{schler2006effects}, recently standardized and truncated to posts between 2–5 sentences by \citet{fisher2024styleremix}. We use this updated version, which includes data from 5 users, each contributing between 700 and 3,000 posts. For each user, we split the data into 80\% training, 10\% validation, and 10\% test sets. The shorter, more focused nature of the posts in this dataset makes it well-suited for analyzing concise writing styles and exploring author-specific patterns. The average word count per post is 40 words, making it the shortest dataset used in this study.

For all mentioned datasets, we perform no pre-processing as the nature of the task requires to work with the raw text containing all stop-words and punctuation. 

\section{Author Verifiers}
To address RQ1, we first need to train authorship verification models. Author verifiers play a crucial role in our study. Our mental model assumes an adversary equipped with a model that verifies whether a specific user, \textit{$user_i$}, is the author of a piece of text. LLM paraphrasing is utilized by the user to change the writing style of the text, aiming to reduce the detection performance of the author verification (AV) model. The ideal outcome of author obfuscation is that the adversary would no longer be able to accurately attribute the text to \textit{$user_i$}.

\subsection{Training Authorship Verifiers}
To train AV models for each user, we train models with two different sets of features. Each feature set attends to different aspects of the text, giving us a comprehensive set of AV models. Both feature sets are widely used in the literature for training AV models and have been proven to be effective for authorship detection.

\textbf{Writeprints}. These are a group of linguistic and syntactic features that have previously been shown to be highly effective for identifying individuals based on the writing style on the internet \cite{abbasi2008writeprints}. Writeprints encompass a wide range of authorship markers, including lexical attributes (e.g., word length distribution, vocabulary richness), syntactic structures (e.g., function word usage, punctuation patterns), and structural aspects (e.g., sentence length variability). In addition, Writeprints incorporates idiosyncratic markers such as character-level variations, misspellings, and special character usage, which help capture an author’s unique stylistic fingerprint. We train two different models using Writeprints, namely logistic regression and XGBoost \cite{chen2016xgboost}, selected for their high interpretability. Our goal is to identify the stronger model to use in the next stage of the study. If logistic regression emerges as the stronger model, we can apply interpretability techniques such as SHAP values to identify the most influential features. On the other hand, if XGBoost performs better, we can leverage the model’s built-in tree-based structure to extract decision-making features directly. Therefore, the choice of interpretability method depends on which model proves to be more effective.

\textbf{Embeddings.} Our second feature set for training AV models are vectorized embeddings. High-dimensional vector embeddings have revolutionized many NLP tasks and have led to significant improvements for many tasks due to their flexibility and representation power. Using embeddings for training author verifiers could help models learn patterns beyond surface-level lexical cues and enable them to make decisions based on semantic similarities too. Hence, we expect the embedding-based AV models to be the most powerful author verifiers in our experiments. However, this comes with a tradeoff. Embeddings-based models, particularly those using complex architectures like BERT \cite{devlin2019bert}, can be harder to interpret than models relying on Writeprint features, where contribution from individual features are more directly observable. We use BERT large (\texttt{bert-large-uncased}) \cite{devlin2019bert} as author verifiers. We set the learning rate to $1e-5$ and the batch size to 8 for training. Each model is trained for 5 epochs, and we save the checkpoint that achieves the best performance on the validation set.

\subsection{Authorship Verification Results}
The training results are shown in Table \ref{tab:training-results}. Comparing the  writeprints-based models with the BERT-based models (columns \textit{original} under both set of features), we observe that BERT-based AVs have a higher F-1 score (0.94 vs. 0.90) than XGBoost and logistic regression on average across all users in all three datasets. This aligns with our expectation that using embeddings as features would result in a stronger AV model. Interestingly, XGBoost and logistic regression models trained with writeprints achieve very close performance (0.90) to BERT-based AV models. These high scores achieved for both sets of features suggests that verifying the author of a text has become a less challenging task for current NLP models. Comparing XGBoost and logistic regression across all users indicates that logistic regression slightly outperforms XGBoost, therefore, for the rest of our analysis in this paper we rely on logistic regression as the model which utilizes writeprints to make predictions.

\begin{table*}[h]
    \small
    \centering
    \renewcommand{\arraystretch}{1.05}
    \begin{tabular}{c|l|ccc|ccc}
        \toprule
        \multirow{2}{*}{\textbf{Dataset}} & \multirow{2}{*}{\textbf{User}} & \multicolumn{3}{c|}{\textbf{Writeprint features}} & \multicolumn{3}{c}{\textbf{Embeddings features}} \\ 
        \cline{3-8}
        & & \textbf{original} & \textbf{LLaMA obf.} & \textbf{GPT obf.} & \textbf{original} & \textbf{LLaMA obf.} & \textbf{GPT-4 obf.} \\ 
        \cline{3-5}
        & & \multicolumn{3}{c|}{\textbf{XGB / LR}} & & & \\

        \hline
        \multirow{10}{*}{\textbf{Yelp}} 
                                & User\_24  & 0.90 / 0.88 & \cellcolor{red!20}0.82 / 0.72 & \cellcolor{red!20}0.83 / 0.72 & 0.89 & \cellcolor{red!20}0.83 & \cellcolor{red!20}0.83\\  
        
                               & User\_13  & 0.87 / 0.89 & 0.44 / 0.53 & 0.36 / 0.60 & 0.90 & \cellcolor{red!20}0.74 & \cellcolor{red!20}0.90\\ 
                               
                                & User\_7  & 0.91 / 0.88 & 0.17 / 0.08 & 0.15 / 0.15 & 0.91 & 0.54 & \cellcolor{red!20}0.72\\ 
                                
                               & User\_9  & 0.89 / 0.87 & 0.48 / 0.55 & 0.23 / 0.63 & 0.94 & 0.73 & \cellcolor{red!20}0.76\\ 
                               
                                & User\_22  & 0.91 / 0.83 & 0.43 / 0.12 & 0.52 / 0.24 & 0.93 & 0.09 & 0.43\\ 
                                
                               & User\_16  & 0.96 / 0.97 & 0.21 / 0.14 & 0.05 / 0.15 & 0.85 & 0.65 & 0.05\\ 
                               
                               & User\_26  & 0.84 / 0.82 & 0.57 / 0.59 & 0.38 / 0.56 & 0.85 & 0.67 & 0.15\\ 
                               
                               &  User\_15  & 0.81 / 0.88 & 0.00 / 0.17 & 0.21 / 0.31 & 0.93 &  0.05 & 0.17\\ 
                               
                               & User\_4  & 0.97 / 0.94 &0.08 / 0.08 & 0.19 / 0.28 & 0.91 &  0.04 & \cellcolor{red!20}0.84 \\ 
                               
                               & User\_6 & 0.88 / 0.84 & 0.57 / 0.66 & 0.00 / 0.56 & 0.85 & 0.63 & 0.14\\ \hdashline
                               
                               & Dataset Avg. &  \textbf{0.89} / 0.88 & \textbf{0.38} / 0.36 & 0.29 / \textbf{0.42} & 0.90 & 0.50 & 0.50\\
        \hline
        \multirow{10}{*}{\textbf{IMDb}} & Hitchcoc  & 0.95 / 0.95 & 0.69 / 0.72 & \cellcolor{red!20}0.84 / 0.76 & 0.98 & \cellcolor{red!20}0.84 & \cellcolor{red!20}0.85\\ 
                               & Boblipton  & 0.92 / 0.91 & 0.69 / 0.68 & \cellcolor{red!20}0.75 / 0.71 &  0.98 & \cellcolor{red!20}0.79 & \cellcolor{red!20}0.80\\ 
                               
                               & SnoopyStyle  & 0.96 / 0.96 & 0.00 / 0.01 & 0.21 / 0.48 & 0.99 & 0.00 & \cellcolor{red!20}0.72\\ 
                               
                               & MartinHafer  & 0.97 / 0.97 & 0.01 / 0.15 & 0.18 / 0.34 & 0.99 & 0.45 & 0.11 \\ 
                               
                               & Bkoganbing  & 0.97 / 0.99 & 0.01 / 0.05 & 0.07 / 0.25 & 0.99 & 0.02 & 0.01\\ 
                               
                               & Horst\_In\_Tr  & 0.97 / 0.99 & 0.01 / 0.19 & 0.25 / 0.18 & 0.98 & 0.20 & 0.53\\ 
                               
                               & Claudio\_carv  & 0.99 / 0.99 & 0.09 / 0.30 & \cellcolor{red!20}0.27 / 0.85 & 1.00 & 0.14 & \cellcolor{red!20}0.93\\
                               
                               & Nogodnomas  & 0.96 / 0.96 & 0.24 / 0.09 & 0.69 / 0.39 & 0.98 & 0.47 & \cellcolor{red!20}0.98\\ 
                               
                               & TheLittleSong  & 0.99 / 0.99 & 0.38 / 0.75 & \cellcolor{red!20}0.80 / 0.96 & 1.00 & 0.76 & \cellcolor{red!20}0.98\\
                               
                               & Leofwine\_dra & 0.96 / 0.97 & 0.70 / 0.71 & \cellcolor{red!20}0.83 / 0.84 & 0.99 & 0.71 & 0.78\\ \hdashline
                               & Dataset Avg. & 0.96 / \textbf{0.97} & 0.28 / \textbf{0.36} & 0.49 / \textbf{0.58} & 0.99 & 0.44 & 0.67\\
        \hline
        \multirow{5}{*}{\textbf{Blog}}  & Blog 5546  & 0.73 / 0.72 & \cellcolor{red!20}0.72 / 0.74 & 0.75 / 0.73 & 0.90 & \cellcolor{red!20}0.81 &\cellcolor{red!20} 0.89\\
        
        & Blog 11518  & 0.86 / 0.81 & \cellcolor{red!20}0.74 / 0.71 & \cellcolor{red!20}0.83 / 0.78 & 0.97 & \cellcolor{red!20}0.82 & \cellcolor{red!20}0.95 \\ 
                               
                               & Blog 25872  & 0.94 / 0.92 & 0.04 / 0.20 & 0.47 / 0.57 & 0.95 & 0.04 & 0.51\\ 
                               
                               & Blog 30102  & 0.76 / 0.76 & \cellcolor{red!20}0.59 / 0.61 & \cellcolor{red!20}0.74 / 0.64 & 0.87 & 0.67 & \cellcolor{red!20}0.85\\ 
                               
                               & blog 30407  & 0.81 / 0.80 & 0.53 / 0.52 & \cellcolor{red!20}0.70 / 0.69 & 0.90 & 0.71 &\cellcolor{red!20}0.84 \\ \hdashline
                               & Dataset Avg. & \textbf{0.82} / 0.80 & 0.52 / \textbf{0.56} & \textbf{0.70} / 0.68 & 0.92 & 0.61 & \cellcolor{red!20}0.81 \\ \hline

         & \textbf{Average} & 0.90 / 0.90 & 0.38 / \textbf{0.41} & 0.45 / \textbf{0.54} & 0.94 & 0.50 & 0.63\\ \bottomrule
    \end{tabular}
    \caption{Reporting F1-score for the user-written class. The tables shows performance of different AV models on the original test set and the paraphrased versions of the test set (LLaMA obf and GPT obf). Red cells indicate a detection performance drop of less than 20\%, suggesting that the obfuscation was not effective.} 
    \label{tab:training-results}
\end{table*}

\subsection{Robustness to Obfuscation}
The main purpose of obfuscation is to evade detection. A robust AV model should be able to identify the real author of an article despite its author being obfuscated. In this section, we examine the robustness of the trained AV models. To assess this, we first need to obtain the paraphrased versions using LLMs. We will refer to this method in the tables as "zero-shot paraphrase", as we are not training the LLMs on the task and we are not providing in context learning examples in the prompt. We prompt the LLMs to paraphrase the text while maintaining its meaning. Here is the prompt template that we use for both LLaMA-3.1 and GPT-4:

\vspace{2mm}

\texttt{Paraphrase the following text to obfuscate the author's identity while maintaining the meaning. Only return the paraphrased text.\\ Input text: \{\} \\ output:}

\vspace{2mm}

The obfuscation results in Table \ref{tab:training-results} seem to reveal a bimodal pattern in obfuscation success across different users (looking at \textit{LLaMA obf} and \textit{GPT obf} columns). Specifically, for both LLMs, there are cases where the classification performance drops significantly, indicating successful obfuscation, but also cases where detection performance remains high (highlighted with red color in the table), suggesting failure to effectively obscure authorship. This variation is particularly evident in the IMDb and Yelp datasets, where for some users (e.g., User\_24, User\_4, and Hitchcoc) LLM paraphrasing causes a very small drop in detection score, whereas for others (e.g., User\_16, User\_15, Bkoganbing) it drops the detection performance significantly. The average scores of the data set also reflect this inconsistency: while LLM paraphrasing leads to overall degradation of detection performance(detection performance goes down from 0.94 to 0.50 and 0.63 for LLaMA and GPT-4 respectively), the variability between users highlights that there is no guarantee of success for every individual. This poses a challenge for practical applications of author obfuscation, as it cannot be universally relied upon for privacy protection.

A key observation from Table \ref{tab:training-results} is that LLaMA-3.1 obfuscation consistently reduces classification accuracy more effectively than GPT-4 across all AV models (BERT, XGBoost and logistic regression). Regardless of whether the classifier is a tree-based model (XGBoost), a linear model (Logistic Regression), or a deep learning-based model (BERT), the LLaMA-obfuscated text is more difficult to attribute to the original author. This finding is crucial because it challenges the common assumption that larger models provide the best obfuscation. Instead, LLaMA-3.1 appears to offer better stylistic transformations for anonymization, leading to a larger drop in detection accuracy. However, since we are not evaluating these two LLMs in terms of their outputs' semantic preservation, we can only conclude that LLaMA-3.1 paraphrasing is more effective than GPT-4 at evading AV detection, which could come at the expense of text quality and semantic similarity to the original text. 

\subsection{Bimodality Testing}
\label{sec:bimodality}
To assess the multimodality of LLM's obfuscation performance, we use Hartigan's dip test \cite{hartigan1985dip}. Hartigan's Dip Test is a statistical test used to assess whether a given distribution is unimodal or multimodal. It measures the maximum difference (or "dip") between the empirical distribution function of the data and the best-fitting unimodal distribution. A higher dip value indicates greater deviation from unimodality. The test produces a p-value, where a small p-value (e.g., < 0.05) suggests that the data is unlikely to be drawn from a unimodal distribution, indicating the presence of multiple modes (e.g., bimodality). In our study, we examine the performance of the obfuscation of each model by calculating the performance drop of each AV model between the original test set and its obfuscated version. For example, in Table \ref{tab:training-results}, the obfuscation performance of GPT-4 on a logistic regression AV for \textit{user 24} in the yelp data set (first row) is $88 - 72 = 0.16$. We apply Hartigan's dip test on the obfuscation performance captured for all users for both author verifiers (logistic regression and BERT) and present the results in Table \ref{tab:dip_test_results}. The results show that LLaMA-3.1 obfuscation exhibits stronger evidence of multimodal behavior for both AV models, while GPT-4 does not exhibit strong evidence of bimodality. This could be because LLaMA-3.1 places greater emphasis on altering writing style, potentially at the expense of content preservation, more so than GPT-4.

\begin{table}[ht]
\centering
\small
\begin{tabular}{l c c}
\toprule
\textbf{Model} & \textbf{GPT-4} & \textbf{LLaMA-3.1} \\
\midrule
Logistic Regression & 0.270 & 0.000 \\
BERT                & 0.572 & 0.050 \\
\bottomrule
\end{tabular}
\caption{Hartigan's Dip Test p-values for GPT-4 and LLaMA-3.1 zero-shot paraphrasing under logistic regression and BERT classifiers. Lower p-values indicate stronger evidence for a bimodal distribution.}
\label{tab:dip_test_results}
\end{table}

\section{Personalized Obfuscation}
Our authorship verification and obfuscation experiments reveal multimodal behavior, meaning that state-of-the-art obfuscation methods perform well for some users but fail for others. To address this multimodality issue in zero-shot paraphrasing, we propose a personalized approach to author obfuscation. Our intuition for this approach is to change the most characteristic features of an author's writing style while paraphrasing rather than apply the same generalized obfuscation approach to all users. To do so, we look at the most important features the trained AV models rely on to make predictions. Having identified the most characteristic features for a given author, we prompt the LLM to paraphrase the text with extra attention to changing that particular stylistic feature. The success of this approach depends on how well we can find the most important feature for each user and how effectively LLMs can change the requested feature.

\subsection{Author-specific Predictive Features with SHAP Values}
\begin{figure}[H]
    \centering
    \includegraphics[width=0.8\linewidth]{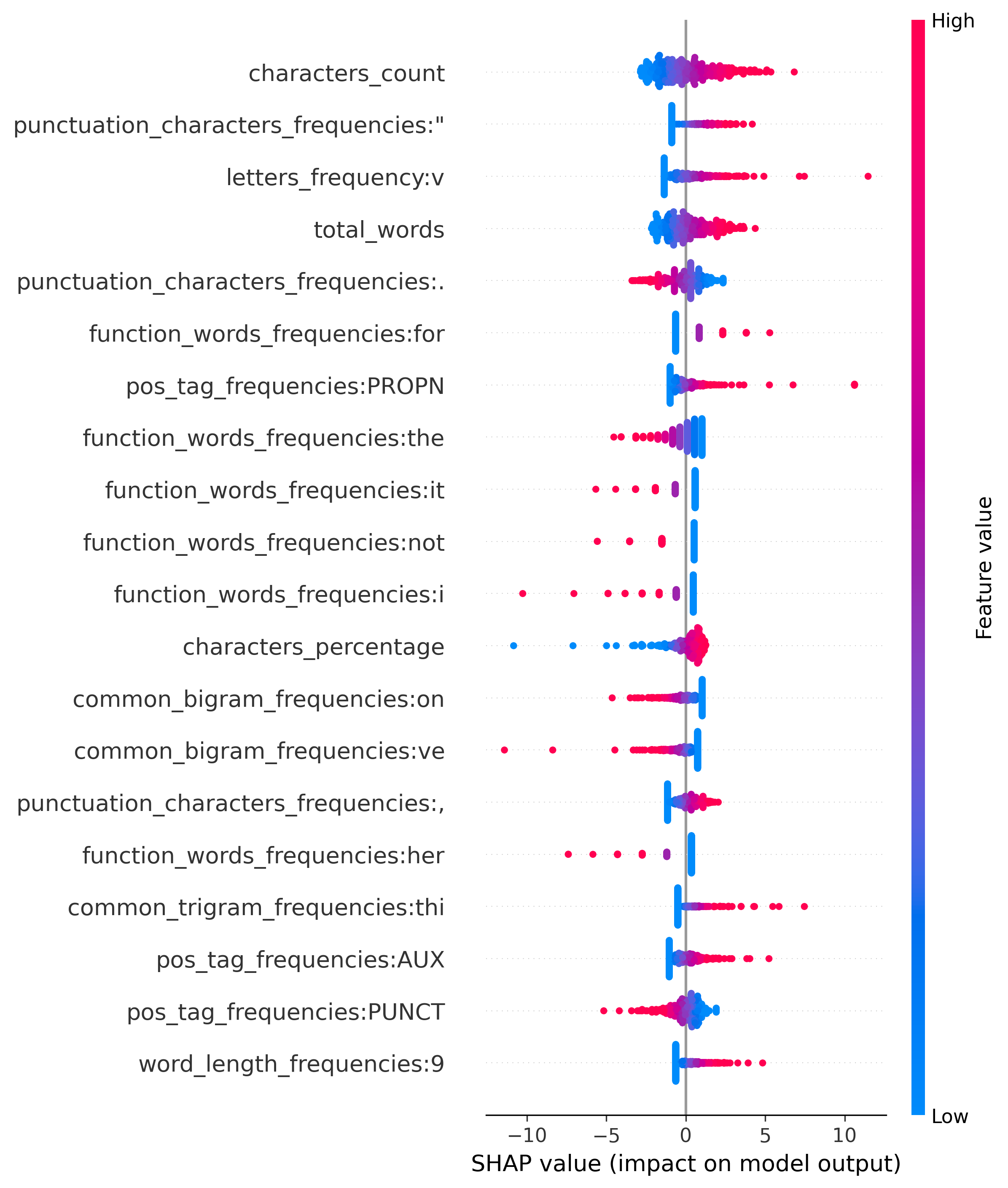}
    \caption{Top features with highest average SHAP values for a given user. The side with higher concentration of red dots indicate the affect of increasing feature value on model's prediction.}
    \label{fig:shap-values}
\end{figure}

To find the features most unique to each author, we use SHAP\cite{hart1989shapley} values. SHAP values, derived from game theory, explain model predictions by quantifying each feature's contribution to the final output, which provides both global and local interpretability. For each author, we found the top features with highest average SHAP values over the validation dataset. This information sheds light on the features that contributed the most to identifying the author in the validation data set. Figure \ref{fig:shap-values} shows the top features and their contributions to model predictions for a particular author. After we learn the top feature with the highest average SHAP value, we use it to generate a  personalized prompt for each author. We first assess the feature's sign and then prompt the model to change the feature accordingly. A negative SHAP value indicates that the corresponding feature has a negative impact on the model's prediction, pushing the prediction away from predicting the user in question as the author of the text. Figure \ref{fig:shap-values} helps us to understand the effect of each feature on the prediction of the model. In the case where increasing the feature value would increase its SHAP value, we design a personalized prompt to decrease that feature's value to confuse the AV model. Here is an example of a prompt designed for a user that has \textit{double quotation mark frequency} as its highest SHAP value feature:

\texttt{Paraphrase the following text to obfuscate the author's identity while maintaining the meaning. Ensure the paraphrased version has more **double quotation marks** than the input. \\ Only return the paraphrased text. \\ Input text: \{\} \\Output: }

\begin{table*}[h]
    \centering
    
    \resizebox{0.8\textwidth}{!}{%
        \begin{tabular}{l| l l c c }
            \toprule
            \textbf{Dataset} & \textbf{Author} & \textbf{Highest SHAP Feature} & \textbf{GPT-4 Personalized obf} & \textbf{Llama Personalized obf}\\
            \midrule
            \multirow{3}{*}{\textbf{Yelp}}  
            & User\_24  & SPACE Pos tag frequency \textcolor{red}{\textdownarrow} & \textcolor{red}{unsuccessful increase} & \textcolor{red}{unsuccessful increase} \\
            & User\_13  & SPACE Pos tag frequency \textcolor{darkgreen}{\textuparrow}   & successful decrease & successful decrease\\
            & User\_7   & Single quotation mark frequency \textcolor{darkgreen}{\textuparrow}  & successful decrease & successful decrease\\
            & User\_9   & Period mark frequency \textcolor{red}{\textdownarrow}  & successful increase & \textcolor{red}{unsuccessful increase} \\
            & User\_22   & SPACE Pos tag frequency \textcolor{darkgreen}{\textuparrow} & successful decrease & successful decrease\\
            & User\_16  & SPACE Pos tag frequency \textcolor{darkgreen}{\textuparrow}  & successful decrease & successful decrease\\
            & User\_26   & Comma frequency \textcolor{red}{\textdownarrow} & successful increase & successful increase \\
            & User\_15   & CCONJ Pos tag frequency \textcolor{red}{\textdownarrow} & successful increase & successful increase \\
            & User\_4   & Dash frequency \textcolor{darkgreen}{\textuparrow} & successful decrease & successful decrease\\
            & User\_6   & SPACE Pos tag frequency \textcolor{darkgreen}{\textuparrow} & successful decrease & successful decrease\\
            \midrule
            \multirow{3}{*}{\textbf{IMDb}}  
            & Hitchcock     & Dash frequency \textcolor{red}{\textdownarrow}  & successful increase & successful increase \\
            & Boblipton     & Comma frequency \textcolor{darkgreen}{\textuparrow}  & successful decrease & successful decrease \\
            & SnoopyStyle    & Period frequency \textcolor{darkgreen}{\textuparrow}  & successful decrease & successful decrease \\
            & MartinHafer    & Exclamation mark frequency \textcolor{darkgreen}{\textuparrow}  & successful decrease  & successful decrease \\
            & Bkoganbing    & PUNCT PoS tag frequency \textcolor{red}{\textdownarrow}  & successful increase & successful increase \\
            & Horst\_In\_Tr    & ADV PoS tag frequency \textcolor{darkgreen}{\textuparrow}  & successful decrease  & successful decrease \\
            & Claudio\_carv    & Function word \textit{"example"} frequency \textcolor{darkgreen}{\textuparrow}  & successful decrease & successful decrease  \\
            & Nogodnomas    & NOUN Pos tag frequency \textcolor{darkgreen}{\textuparrow}  & \textcolor{red}{unsuccessful decrease} & \textcolor{red}{unsuccessful decrease} \\
            & TheLittleSong    & Single quotation mark frequency \textcolor{darkgreen}{\textuparrow}  & successful decrease & successful decrease \\
            & Leofwine\_dra    & Uppercase characters percentage \textcolor{darkgreen}{\textuparrow}  & successful decrease & successful decrease \\
            \midrule
            \multirow{3}{*}{\textbf{Blog}}  
            & Blog\_5546  & characters count \textcolor{darkgreen}{\textuparrow}   & successful decrease & successful decrease \\\
            & Blog\_11518 & characters percentage \textcolor{darkgreen}{\textuparrow}  & \textcolor{red}{unsuccessful decrease} & \textcolor{red}{unsuccessful decrease} \ \\
            & Blog\_25872 & Period mark frequency \textcolor{darkgreen}{\textuparrow}  & successful decrease& successful decrease \\
            & Blog\_30102 & Question mark frequency \textcolor{darkgreen}{\textuparrow}  & successful decrease & successful decrease\ \\
            & Blog\_30407 & Double quotation mark frequency \textcolor{red}{\textdownarrow} & successful increase & successful increase\ \\
            
            \bottomrule
        \end{tabular}
        }
        
    \caption{Top SHAP Feature Per Author and its successful or unsuccessful change in obfuscated text, for both GPT-4 and Llama. }
    \label{tab:feature_change_success}
\end{table*}

\subsection{Personalized Obfuscation Results}

After designing personalized prompts for all users, we prompt both LLMs with the personalized prompts. To evaluate the obfuscation performance, we use the same AV models trained previously. As in the previous setting, a bigger drop in the AV model's detection performance indicates a better obfuscation performance. To ensure generalizability, we evaluate the obfuscation performance using both BERT and logistic regression AV models. We first evaluate the success of LLMs in effectively implementing the personalized obfuscation prompts, and then evaluate the performance of the personalized obfuscation against both AV models.

\vspace{1mm}

\textbf{How well can LLMs change a specific feature while paraphrasing text?}
The success of our proposed method relies on how effectively LLMs can modify a requested feature, $f_i$. To evaluate this, we compare the value of $f_i$ in the original text to its value in the paraphrased version generated with the personalized prompt. The results, shown in Table \ref{tab:feature_change_success}, indicate that both LLMs can generally adjust the requested feature successfully. Yet, both models struggle to decrease noun frequency and increase whitespace frequency, while showing strong ability in adjusting punctuation marks, uppercase letters, adverbs, and other stylistic elements.

\vspace{1mm}
\textbf{How does personalized obfuscation evade detection by a logistic regression model as AV?} As shown in Table \ref{tab:lr_personalized_obf}, personalized obfuscation with GPT-4, consistently reduces the AV detection performance across all datasets, indicating a more effective method than zero-shot prompting. This can be seen by comparing GPT-4’s personalized obfuscation scores with its zero-shot paraphrasing scores. For example, in the Yelp dataset, the AV model achieves an f-1 score of 0.42 on a zero-shot paraphrased text, but this drops to 0.40 for personalized obfuscation. A similar pattern is observed in the IMDb dataset, where the F1 score decreases from 0.58 in the zero-shot setting to 0.51 in the personalized setting, and in the Blog dataset, where it drops from 0.68 to 0.55. This suggests that personalized obfuscation introduces more targeted changes to the writing style, which makes it harder for the author verifier to detect the original author.

For LLaMA-3.1, personalized obfuscation reduces AV detection performance compared to zero-shot paraphrasing in the Yelp and Blog datasets but shows mixed results in the IMDb dataset. In the Yelp dataset, the average author verification score decreases from 0.36 in the zero-shot setting to 0.35 with personalized obfuscation, indicating that LLaMA’s personalized outputs make it harder for the classifier to identify the original author. Similarly, in the Blog dataset, the score drops from 56\% to 52\%, reflecting more effective obfuscation through personalization. However, in the IMDb dataset, the score increases from 36\% to 39\%, suggesting that LLaMA’s personalized obfuscation is less effective in this domain, potentially due to stronger stylistic consistency in the underlying text. These findings highlight that LLaMA’s personalized obfuscation is more successful in less structured domains like Yelp and Blog, while it struggles to evade detection in more formal datasets like IMDb.

\vspace{2mm}

\textbf{How does personalized obfuscation evade detection by a BERT model as AV?}

Our personalized prompting method was designed to be most effective for a logistic regression author verifier or any other machine learning model which works with writeprint/lexical features. This is because our approach relies on the most important feature, which could be mostly identified in simpler models that don't rely on vectorized embeddings. In this section we investigate the efficacy of our obfuscation approach for a BERT classifier. Our results in Table \ref{tab:bert_personalized_obfs} show that our method works for a BERT classifier too. This result indicates that providing informative details about the obfuscation process in the prompt could be beneficial regardless of the author verifier (writeprint-based or embedding-based).  

As shown in Table \ref{tab:bert_personalized_obfs}, personalized obfuscation improves obfuscation performance (i.e., leads to a greater drop in classification accuracy) compared to zero-shot paraphrasing for both LLMs, particularly in the Yelp and IMDb datasets. In the Yelp dataset, BERT's average author verification F1 score decreases from 0.50 in the zero-shot setting to 0.48 with personalized obfuscation for GPT-4 and from 0.50 to 0.40 for LLaMA-3.1, indicating that LLaMA-3.1 benefits more from personalization. A similar trend is observed in the IMDb dataset, where the average verification score drops from 0.67 to 0.61 for GPT-4 and from 0.40 to 0.37 for LLaMA-3.1. However, this pattern does not hold for the Blog dataset, where personalized obfuscation does not produce a consistent drop in verification performance. This could be due to the shorter text lengths in the Blog dataset, which may limit the impact of style-based obfuscation.

\begin{table}[h]
\centering
\resizebox{\columnwidth}{!}{%
\begin{tabular}{l |  l | c |  c c c c}
\toprule
\textbf{Dataset} & \textbf{User} & \multicolumn{1}{c}{\textbf{Test Set og}} & \multicolumn{2}{c}{\textbf{Zero-Shot Paraphrase}} & \multicolumn{2}{c}{\textbf{Personalized Obfs}} \\
   \cmidrule(lr){4-5} \cmidrule(lr){6-7}
& & & \textbf{GPT-4} & \textbf{LLaMA} & \textbf{GPT-4} & \textbf{LLaMA} \\
\midrule
\multirow{10}{*}{\textbf{Yelp}} 
& User\_24 & 0.88 & 0.72 & 0.72 & 0.71 & 0.74 \\
& User\_13 & 0.89 & 0.60 & 0.53 & 0.58 & 0.51 \\
& User\_7 & 0.88  & 0.15 & 0.08 & 0.18 & 0.08 \\
& User\_9 & 0.87  & 0.63 & 0.55 & 0.61 & 0.53 \\
& User\_22 & 0.83 & 0.24 & 0.12 & 0.14 & 0.37 \\
& User\_16 & 0.97 & 0.05 & 0.14 & 0.07 & 0.21 \\
& User\_26 & 0.82 & 0.15 & 0.59 & 0.57 & 0.52 \\
& User\_15 & 0.88 & 0.17 & 0.17 & 0.29 & 0.05 \\
& User\_4 & 0.94  & 0.84 & 0.08 & 0.29 & 0.01 \\
& User\_6 & 0.84  & 0.14 & 0.66 & 0.54 & 0.45 \\
\hdashline
& \textbf{Average} & 0.88 & 0.42 & 0.36 & 0.40 & 0.35 \\
\midrule
\multirow{10}{*}{\textbf{IMDb}} 
& Hitchcoc & 0.98 & 0.85 & 0.72 & 0.75 & 0.64 \\
& Boblipton & 0.98 & 0.80 & 0.68 & 0.66 & 0.65 \\
& SnoopyStyle & 0.99 & 0.72 & 0.01 & 0.11 & 0.18 \\
& MartinHafer & 0.99 & 0.11 & 0.15 & 0.31 & 0.12 \\
& Bkoganbing & 0.99 & 0.01 & 0.05 & 0.11 & 0.11 \\
& Horst\_In\_Tr & 0.98 & 0.53 & 0.19 & 0.22 & 0.16 \\
& Claudio\_carv & 1.00 & 0.93 & 0.30 & 0.81 & 0.41 \\
& Nogodomas & 0.98 & 0.98 & 0.09 & 0.38 & 0.14 \\
& TheLittleSong & 1.00 & 0.98 & 0.75 & 0.97 & 0.76 \\
& Leofwine\_dra & 0.99 & 0.78 & 0.71 & 0.80 & 0.73 \\
\hdashline
& \textbf{Average} & 0.97 & 0.58 & 0.36 & 0.51 & 0.39 \\
\midrule
\multirow{5}{*}{\textbf{Blog}} 
& Blog\_5546 & 0.72  & 0.73 & 0.74 & 0.70 & 0.72 \\
& Blog\_11518 & 0.81 & 0.78 & 0.71 & 0.78 & 0.73 \\
& Blog\_25872 & 0.92 & 0.57 & 0.20 & 0.25 & 0.25 \\
& Blog\_30102 & 0.76 & 0.64 & 0.61 & 0.69 & 0.61 \\
& Blog\_30407 & 0.80 & 0.69 & 0.52 & 0.35 & 0.29 \\
\hdashline
& \textbf{Average} & 0.80 & 0.68 & 0.56 & 0.55 & 0.52\\
\bottomrule
\end{tabular}}
\caption{Comparison between personalized obfuscation with LLMs vs. zero-shot obfuscation on logistic regression AV model across different datasets and users.}
\label{tab:lr_personalized_obf}
\end{table}

\begin{table}[h]
\centering
\resizebox{\columnwidth}{!}{%
\begin{tabular}{l |  l | c  c c c c}
\toprule
\textbf{Dataset} & \textbf{User} & \multicolumn{1}{c}{\textbf{Test og}} & \multicolumn{2}{c}{\textbf{Zero-Shot Paraphrase}} & \multicolumn{2}{c}{\textbf{Personalized Obfs}} \\
   \cmidrule(lr){4-5} \cmidrule(lr){6-7}
& & & \textbf{GPT-4} & \textbf{LLaMA} & \textbf{GPT-4} & \textbf{LLaMA} \\
\midrule
\multirow{10}{*}{\textbf{Yelp}} 
& User\_24 & 0.89 & 0.83 & 0.83 & 0.82 & 0.76 \\
& User\_13 & 0.90 & 0.90 & 0.74 & 0.82 & 0.88 \\
& User\_7 & 0.91  & 0.72 & 0.54 & 0.78 & 0.73 \\
& User\_9 & 0.94  & 0.76 & 0.73 & 0.81 & 0.84 \\
& User\_22 & 0.93 & 0.43 & 0.09 & 0.44 & 0.49 \\
& User\_16 & 0.85 & 0.05 & 0.65 & 0.10 & 0.08 \\
& User\_26 & 0.85 & 0.15 & 0.67 & 0.12 & 0.04 \\
& User\_15 & 0.93 & 0.17 & 0.05 & 0.15 & 0.00 \\
& User\_4 & 0.91  & 0.84 & 0.04 & 0.61 & 0.00 \\
& User\_6 & 0.85  & 0.14 & 0.63 & 0.19 & 0.22 \\
\hdashline
& \textbf{Average} & 0.90 & 0.50 & 0.50 & 0.48 & 0.40 \\
\midrule
\multirow{10}{*}{\textbf{IMDb}} 
& Hitchcoc & 0.98 & 0.85 & 0.84 & 0.89 & 0.42 \\
& Boblipton & 0.98 & 0.80 & 0.79 & 0.70 & 0.67 \\
& SnoopyStyle & 0.99 & 0.72 & 0.00 & 0.28 & 0.37 \\
& MartinHafer & 0.99 & 0.11 & 0.45 & 0.03 & 0.02 \\
& Bkoganbing & 0.99 & 0.01 & 0.02 & 0.00 & 0.00 \\
& Horst\_In\_Tr & 0.98 & 0.53 & 0.20 & 0.57 & 0.23 \\
& Claudio\_carv & 1.00 & 0.93 & 0.14 & 0.92 & 0.14 \\
& Nogodomas & 0.98 & 0.98 & 0.09 & 0.98 & 0.52 \\
& TheLittleSong & 1.00 & 0.98 & 0.76 & 1.00 & 0.94 \\
& Leofwine\_dra & 0.99 & 0.78 & 0.71 & 0.71 & 0.71 \\
\hdashline
& \textbf{Average} & 0.99 & 0.67 & 0.40 & 0.61 & 0.37 \\
\midrule
\multirow{5}{*}{\textbf{Blog}} 
& Blog\_5546 & 0.90  & 0.89 & 0.81 & 0.91 & 0.87 \\
& Blog\_11518 & 0.97 & 0.95 & 0.82 & 0.96 & 0.83 \\
& Blog\_25872 & 0.95 & 0.51 & 0.04 & 0.32 & 0.32 \\
& Blog\_30102 & 0.87 & 0.85 & 0.67 & 0.85 & 0.70 \\
& Blog\_30407 & 0.90 & 0.84 & 0.71 & 0.79 & 0.42 \\
\hdashline
& \textbf{Average} & 0.92 & 0.71 & 0.61 & 0.77 & 0.63 \\
\bottomrule
\end{tabular}}
\caption{Performance comparison between personalized obfuscation with LLMs vs. zero-shot obfuscation on BERT AV model across different datasets and users}
\label{tab:bert_personalized_obfs}
\end{table}

\vspace{2mm}

\textbf{How does personalized obfuscation affect obfuscation performance multi-modality?}
To evaluate whether our proposed method mitigates the multi-modality issue discussed in Section \ref{sec:bimodality}, we apply Hartigan's Dip Test to the personalized obfuscation results in Tables 4 and 5. The results, shown in Table \ref{tab:dip_test_results_personalized}, indicate that none of the models exhibit a p-value below 0.05, suggesting that the new distributions are less likely to follow a multi-modal pattern. This implies that personalized obfuscation helps reduce the variability in obfuscation performance across different users, leading to more consistent results.

\begin{table}
\centering
\small
\begin{tabular}{l c c}
\toprule
\textbf{Model} & \textbf{GPT-4} & \textbf{LLaMA-3.1} \\
\midrule
Logistic Regression & 0.061 & 0.081 \\
BERT                & 0.407 & 0.429 \\
\bottomrule
\end{tabular}
\caption{Hartigan's Dip Test p-values for GPT-4 and LLaMA-3.1 personalized obfuscation under logistic regression and BERT classifiers. Lower p-values indicate stronger evidence for a bimodal distribution.}
\label{tab:dip_test_results_personalized}
\end{table}

\section{Conclusion}
Our study demonstrates that simply prompting large language models (LLMs) to obfuscate the author leads to a noticeable drop in author verification (AV) performance. However, our user-wise analysis reveals a bimodal distribution in obfuscation effectiveness. while the average drop in performance is substantial, for some authors the drop is relatively small, whereas for others it is significantly larger. This highlights that a one-size-fits-all approach to obfuscation may not work equally well across different writing styles.

By analyzing SHAP values, we identified the most influential features unique to each author’s writing style. These features represent stylistic patterns that are particularly useful for author verification and could be targeted more effectively in obfuscation strategies. Our personalized obfuscation method, which leverages author-specific SHAP values, helps mitigate this bimodality by adapting the obfuscation process to each author’s unique writing style. This targeted approach further improves obfuscation effectiveness and makes it more difficult for author verification models to attribute text to the original author.

\section*{Limitations}
Our study has some limitations that should be considered when interpreting the results. First, our approach is evaluated on a relatively small set of 25 authors, which may limit the generalizability of the findings. However, it is important to note that these 25 authors are drawn from three distinct datasets (Yelp, IMDb, and Blog), which helps to introduce some diversity in writing styles and domains. This cross-domain evaluation increases the robustness of our findings despite the small sample size. Furthermore, other significant works in authorship obfuscation have used a similar number of users. For instance, the DP-Prompt paper \cite{utpala2023locally} applied their methodology on 10 users from the Yelp dataset and 10 users from the IMDb dataset, totaling 20 users, and AUTHORMIX dataset by \citet{fisher2024styleremix} has 30k posts by only 14 users. This suggests that evaluating authorship obfuscation on a similar scale is consistent with established practices in the field.

Second, this paper proposes a personalized prompting approach for authorship obfuscation that relies on modifying a single highly predictive feature using a single prompting strategy. Despite only using one feature and prompt structure, we were still able to observe that personalization is effective in improving obfuscation performance. Fine-tuning and optimizing prompts could potentially improve obfuscation performance, especially in more complex linguistic settings. Expanding the scope of prompt design and feature combination remains an important area for future work, as it could provide deeper insights into the stylistic patterns that enable successful obfuscation.

\bibliography{custom}

\end{document}